\documentclass[sigconf,nonacm,10pt]{acmart}

\AtBeginDocument{%
  }

\acmConference[Anonymous Workshop Submission]{}{}{}
\acmYear{2026}

\usepackage{graphicx}
\graphicspath{{figures/}}
\usepackage{booktabs}
\usepackage{amsmath}
\usepackage{xspace}
\usepackage{multirow}
\usepackage{xcolor}
\usepackage{balance}
\usepackage{tikz}
\usepackage{pifont}
\usepackage{makecell}
\usepackage{graphicx}
\usepackage{subcaption} 
\usepackage{enumitem}

\usetikzlibrary{positioning,arrows.meta,calc,backgrounds,fit,shapes.geometric,decorations.pathreplacing}

\newcommand{\phisym}{\ensuremath{\varphi}\xspace}
\newcommand{\phiwide}{\phisym}

\newcommand{\sys}{\textsc{CacheScout}\xspace}
\newcommand{\kvflow}{KVFlow\xspace}
\newcommand{\continuum}{Continuum\xspace}
\newcommand{\vllm}{vLLM\xspace}

\usepackage{amsthm}
\usepackage{algorithm}
\usepackage{algpseudocode}

\begin{document}

\title{A Policy-Driven Runtime Layer for Agentic LLM Serving}


\author{Rui Zhang}
\affiliation{%
  \institution{University of California, Santa Cruz}
  \country{}}

\author{Chaeeun Kim}
\affiliation{%
  \institution{University of California, Santa Cruz}
  \country{}}

\author{Liting Hu}
\affiliation{%
  \institution{University of California, Santa Cruz}
  \country{}}

\renewcommand{\shortauthors}{Zhang et al.}

\begin{abstract}
Multi-agent LLM systems have become the dominant production
workload, but the serving stack was not built for them. The agent
framework above knows agent identities, role schemas, and dispatch
structure but never sees an engine-level event; the serving engine
below sees every event but knows nothing about agents. A surprising
number of cross-cutting policies depend on both: prefix caching,
batch shaping, speculative execution, fairness, tool-result
memoization, safety enforcement, and more. Each lives
in the seam between the two layers and is currently solved by a
one-off patch into one neighbor or the other.

We argue this seam is best addressed by an architectural change
rather than point fixes: insert a third tier, an \textbf{agent
runtime layer}, between the framework and the engine, exposing
four primitives (\texttt{observe}, \texttt{score}, \texttt{predict},
\texttt{act}) into which any agent-aware policy plugs, with
\emph{agent identity} as the shared coordinate. We map nine
concrete policies onto the layer and validate the abstraction in
depth on the one with the largest immediate serving-cost lever:
KV caching across sessions, instantiated as \sys, which learns the
per-workload agent transition matrix online and uses it for
survival-based eviction and between-step prefetch. Preliminary
results on five real multi-agent workloads show +13 to +37\,pp
cache hit-rate lift, 12\% to 29\% lower mean TTFT, and 6\% to
14\% higher throughput over an unmodified serving stack.

\end{abstract}



\maketitle

\section{Introduction}
\label{sec:intro}

\textit{Multi-agent LLM workloads dominate production.}
A multi-agent LLM system decomposes a single user request into a
coordinated trajectory of model calls, for example a supervisor
that routes work to specialists, a hierarchical organization that
recursively breaks down objectives, or a swarm of peers that hand
off through tool returns. The pattern is prevalent in production. 
A majority of organizations now run agentic systems in
production~\cite{langchain2026soa}, and framework adoption nearly
doubled year-over-year~\cite{datadog2026soa}. The multi-agent
regime is no longer a research curiosity; it is the workload the
serving stack has to serve.

\begin{figure}[t]
\centering
\includegraphics[width=\columnwidth]{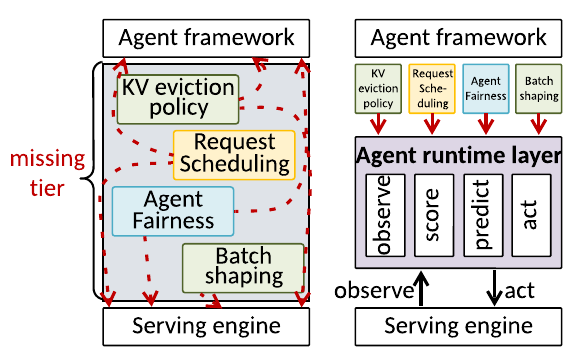}
\caption{The seam between agent framework and serving engine.
\textbf{Today (left):} cross-cutting policies have no shared home and are
patched into one neighbor or the other.
\textbf{Proposed (right):} an agent runtime layer hosts these
policies on four primitives with agent identity as the shared
coordinate.}
\Description{Two side-by-side panels. Left panel shows a two-tier
stack with the agent framework on top and the serving engine on
the bottom; floating between them are four policy bubbles (KV
cache, batch shape, fairness, safety) connected by dashed arrows
to one of the two tiers. Right panel shows the same stack with a
new middle tier called Agent Runtime Layer hosting the four
primitives (observe, score, predict, act); the four policy bubbles
are now connected to this middle tier by solid arrows.}
\label{fig:gap}
\end{figure}

\textit{Many cross-cutting policies live between the framework
and the engine.}
Production multi-agent systems sit on a two-tier stack today
(Figure~\ref{fig:gap}, left): an
agent framework above (AutoGen, LangGraph, MetaGPT, swarm-style
peers) that emits LLM API calls under its own dispatch logic, and
a serving engine below (vLLM~\cite{kwon2023vllm},
SGLang~\cite{zheng2024sglang}) that prefills KV, batches requests,
and returns tokens. A surprising number of production-relevant policy decisions live
\emph{between} them. For example:
\begin{itemize}[leftmargin=*, itemsep=0.2em]
\item Deciding which KV blocks recur across sessions and should
not be evicted is a question about agent identity, not block
hashes.
\item Deciding which incoming requests likely share state and
should be batched together is a question about agent dispatch,
not queue order.
\item Deciding which agent class deserves a larger share of
capacity is a question about agent fairness, not per-request
priority.
\end{itemize}
Table~\ref{tab:policies} sketches nine such policies, from KV
caching and batch shaping to speculative agent execution and
fairness. Each is currently treated by a one-off patch into
either the framework (a wrapper that hand-codes a workflow graph
or a retry decorator) or the engine (a TTL pin, a scheduler
heuristic, a flag that biases eviction). The recurring shape is
the same in every case: the policy needs agent-level metadata,
the policy-relevant events arrive at the engine, and neither
neighbor alone holds both.

\textit{The gap is architectural.}
The framework knows agent identities, roles, and dispatch but
never sees an engine event; the engine sees every event but knows
nothing about agents. The policies above all need the
\emph{intersection}: events tagged by agent identity, accumulated
across sessions, queryable on the engine hot path. That
intersection cannot be retrofitted cleanly into either neighbor.
The right response is architectural: \emph{insert a third tier}
that owns the intersection and exposes it through a stable
interface.

\textit{An agent runtime layer with four primitives.}
We propose the missing tier as an \emph{agent runtime layer}
(Figure~\ref{fig:gap}, right): a substrate between framework and
engine that owns the cross-cutting agent metadata neither neighbor
has, exposing four primitives any agent-aware policy plugs into: \texttt{observe} (receive events
from a neighbor), \texttt{score} (return per-item priority for a
neighbor's ranking decision), \texttt{predict} (return a
probabilistic forecast over future agent activity), and
\texttt{act} (issue an off-critical-path side-effect). Together
they form an
observe\,$\to$\,score\,$\to$\,predict\,$\to$\,act feedback loop,
with \emph{agent identity} as the shared coordinate across all
four. The primitives are policy-agnostic; every policy in
Table~\ref{tab:policies} maps onto them
(\S\ref{sec:layer:primitives} formalizes the contract).

\textit{Case study: an agent-aware KV cache.}
We validate the abstraction in depth on KV caching, the policy
with the largest immediate serving-cost lever. Multi-agent prompts
carry a recurring system+tools \emph{agent anchor} (0.34 to 0.52
of every first-turn prompt) that reactive prefix caches drop to
LRU before the next session arrives. \sys closes that gap by
learning the per-workload first-order Markov agent transition
matrix online and using it for survival-based eviction
(\texttt{score}) and between-step prefetch
(\texttt{predict}+\texttt{act}), with bounded state
($\le$\,20\,KB/workload) and graceful LRU fallback.

\textit{Our contributions are as follows:}
\begin{itemize}[leftmargin=*]
\item \textbf{The agent runtime layer.} The third tier the serving
stack is missing, with a four-primitive interface that any
agent-aware policy plugs into, and nine concrete policies mapped
onto it (Table~\ref{tab:policies}).
\item \textbf{\sys: an instantiation for KV caching.} The four
primitives realized in three components: transition learner,
survival-probability scorer, and workload-gated cross-session
prefetch.
\item \textbf{Preliminary evidence the gap matters.} On five
multi-agent workloads, \sys lifts hit rate by +13 to +37\,pp,
cuts mean TTFT by 12\% to 29\%, and raises throughput by 6\% to
14\%.
\end{itemize}
\section{The Agent Runtime Layer}
\label{sec:layer}

\subsection{Architecture and Four Primitives}
\label{sec:layer:primitives}

The agent runtime layer is a third tier between the framework
above and the serving engine below that owns the cross-cutting
agent metadata neither neighbor has. The framework keeps emitting
LLM API calls; the engine keeps serving them; the runtime makes
the agent-level policy decisions both neighbors lack the metadata
to make.

The runtime's contract with its neighbors reduces to four generic
primitives, each parameterized by the policy hosted on the layer:
\begin{itemize}[leftmargin=*]
\item \texttt{observe(event)}: the runtime is notified of a
policy-relevant event from a neighbor (a block touch, a request
arrival, a tool return, a tenant deadline). It updates internal
state tagged by \emph{agent identity}, the shared coordinate the
layer adds on top.
\item \texttt{score(item)}: the runtime returns a per-item
priority a neighbor uses to rank a decision (which block to
evict, which request to admit, which speculative branch to
commit, which tenant to throttle).
\item \texttt{predict(horizon)}: the runtime returns a
probabilistic forecast over future agent activity (next agent,
next tool, next tenant burst). The forecast's form is
policy-specific; the primitive is not.
\item \texttt{act(side\_effect)}: the runtime issues an
off-critical-path side-effect to a neighbor (warm a cache,
pre-load weights, hedge a speculative call, defer a request).
\texttt{act} is the only primitive that mutates neighbor state.
\end{itemize}
Together they form an observe\,$\to$\,score\,$\to$\,predict\,$\to$\,act
feedback loop, with agent identity as the shared coordinate that
unifies events, scores, predictions, and actions across policies.

\subsection{Policies the Layer Can Host}
\label{sec:layer:policies}

KV caching is the policy we validate in depth (\S\ref{sec:casestudy}),
but the layer is a substrate for any agent-aware policy in the
same seam. Table~\ref{tab:policies} sketches nine concrete
instances grouped into three families and shows how each maps
uniformly onto the four primitives. The first row of the
memory-and-state group is the case study we develop; the other
eight rows are policies currently solved by ad-hoc patches into
one neighbor that, we argue, belong on the layer for the same
architectural reason.

We do not claim every cell is a finished design; the contribution
of the table is that the same four-primitive shape recurs across
otherwise unrelated problems, all of which need agent identity and
none of which fit cleanly on either neighbor alone. The other
eight rows remain open challenges for future work.

\begin{table*}[t]
    \centering
    \small
    \caption{Nine agent-aware policies the runtime layer can host,
    mapped onto the four primitives.}
    \label{tab:policies}
    \setlength{\tabcolsep}{4pt}
    \begin{tabular}{l l l l l}
    \toprule
    \textbf{Policy} & \texttt{observe} & \texttt{score} & \texttt{predict} & \texttt{act} \\
    \midrule
    \multicolumn{5}{l}{\textbf{Memory and state}} \\
    KV cache (\textbf{this paper})        & block touches            & eviction priority    & next agent             & warm next anchor \\
    Tool-result memoization      & tool args / result       & memo value           & next tool call         & prefetch / invalidate \\
    \midrule
    \multicolumn{5}{l}{\textbf{Compute scheduling}} \\
    Batch shaping / admission    & queue, prefill ratios    & batch priority    & agent composition              & pack/defer/hedge request \\
    Disagg.\ P/D placement       & per-agent token ratios   & placement            & P/D workload mix        & route by P/D ratio \\
    Speculative agent execution  & transition history       & branch commit        & next agent + input     & issue spec.\ call \\
    Inference adapters/drafters    & per-agent adapter quality & adapter fit  &  agent $\to$ adapter set & apply adapter/drafter \\
    \midrule
    \multicolumn{5}{l}{\textbf{Operational policies}} \\
    Fairness                     & agent resource usage     & long-term share per agent &  agent demand & adjust resource share \\
    Safety / guardrails          & agent outputs            & risk per call        & likely violator        & block / rewrite / escalate \\
    Failure hedging / retry      & failure traces           & retry priority       & likely failure         & warm fallback / hedge \\
    \bottomrule
    \end{tabular}
\end{table*}

\subsection{Properties}
\label{sec:layer:properties}

Three properties make the layer useful independent of any one
policy. \emph{Framework-agnostic}: it requires only LLM API calls
carrying metadata, a contract satisfied by AutoGen,
LangGraph, MetaGPT, ReAct, and swarm frameworks.
\emph{Engine-portable}: each hosted policy needs only the engine
hooks its primitives use, not engine internals; the KV-cache case
study uses exactly two hooks (an \texttt{observe} forward on
block touches and a \texttt{score} consult on eviction).
\emph{Policy-modular}: distinct policies share the layer's
\texttt{observe} stream but compose against the same four
primitives, so adding a new policy does not require new engine
surgery.

\section{Case Study: Agent-Aware KV Cache}
\label{sec:casestudy}

\begin{figure}[t]
    \centering
    \includegraphics[width=\columnwidth]{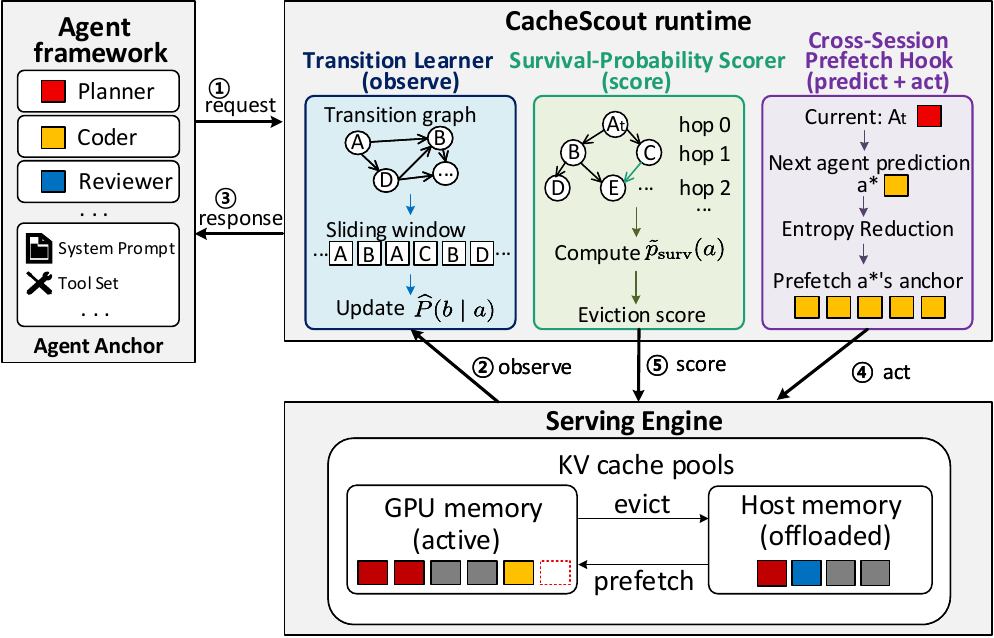}
    \caption{\sys instantiates the runtime layer for KV caching with
    three components: a transition learner (\texttt{observe}), a
    survival-probability scorer (\texttt{score}), and a cross-session
    prefetch hook (\texttt{predict} + \texttt{act}).}
    \Description{Architectural diagram of CacheScout as a runtime
    layer between the agent framework and the serving engine. Shows
    the three internal components (transition learner, scorer,
    prefetch hook) and the two engine hooks they connect to.}
    \label{fig:arch}
  \end{figure}

We instantiate the layer on KV caching, the policy with the
largest immediate serving-cost lever (Figure~\ref{fig:arch}).
\S\ref{sec:casestudy:obs} gives the three workload observations
that motivate the design, and \S\ref{sec:casestudy:learn} through
\S\ref{sec:casestudy:prefetch} realize the four primitives in
three components.

\subsection{Workload Observations}
\label{sec:casestudy:obs}
\label{sec:motivation}

We characterize five real multi-agent workloads
(MMLU~\cite{hendrycks2020measuring},
MT-Bench~\cite{zheng2023judging},
GAIA~\cite{mialon2024gaia},
GSM8K~\cite{cobbe2021training},
HumanEval~\cite{chen2021evaluating}), each running on the same
six-agent
supervisor framework with
Llama-3.1-8B-Instruct~\cite{grattafiori2024llama} and AutoGen~\cite{wu2024autogen} SelectorGroupChat dynamic
LLM-selected routing. Three observations shape the design.

\begin{figure}[t]
  \centering
  \begin{subfigure}[b]{0.48\columnwidth}
    \includegraphics[width=\textwidth]{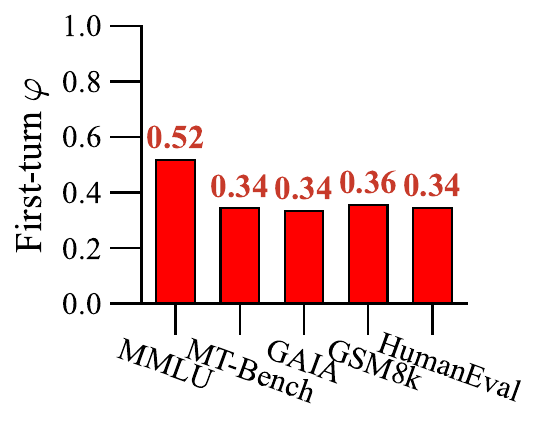}
    \caption{First-turn $\phiwide$.}
    \label{fig:m_phi_first_turn}
  \end{subfigure}
  \hfill
  \begin{subfigure}[b]{0.48\columnwidth}
    \includegraphics[width=\textwidth]{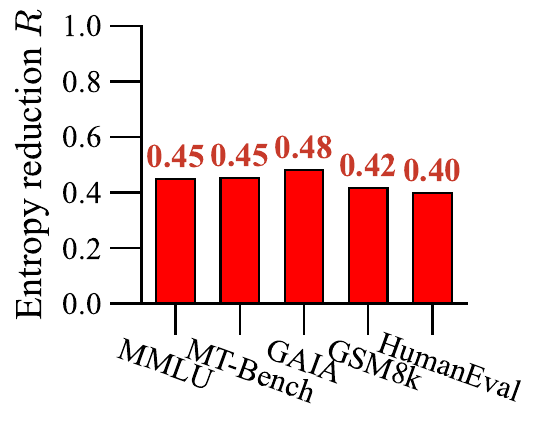}
    \caption{Entropy Reduction R}
    \label{fig:m_entropy_reduction}
  \end{subfigure}
  \caption{Cross-session structure on five multi-agent workloads.
  \textbf{(a)} First-turn shareable surface $\phiwide$.
  \textbf{(b)} Relative conditional entropy reduction~$R$.}
  \Description{Two side-by-side bar charts across five workloads.
  Left: first-turn shareable surface phi, all between 0.34 and 0.52.
  Right: relative conditional entropy reduction R, all between 0.40
  and 0.48.}
  \label{fig:motivation_locality}
\end{figure}

\paragraph{\textbf{Observation 1:} a non-trivial cross-session surface,
concentrated at session start.}
\label{sec:motivation:phi}
\label{sec:motivation:phi_decay}
Let
$\phiwide = (\text{system}+\text{tools}+\text{few-shots}) / \break
(\text{total prompt tokens})$
be the share of each prompt that forms the agent anchor and
recurs identically across sessions. $\phiwide$ upper-bounds the
surface a cross-session cache can amortize; the residual is
per-session history no cross-session layer can reuse.
Figure~\ref{fig:motivation_locality}(a) reports first-turn $\phiwide$: the
surface is non-trivial on every workload (0.34 to 0.52). The
surface does \emph{not} persist within a session. Anchor bytes
stay fixed while per-session history grows linearly with turn
count, so per-turn $\phiwide$ decays monotonically with depth.
This exposes two regimes: \emph{within-session} reuse (already
captured by reactive block-level prefix caching) and
\emph{across-session} reuse, where the anchor is the recurring
substrate but reactive caches retain it only by accident. The
latter is where the case study adds value.

\paragraph{\textbf{Observation 2:} the next agent is substantially
predictable.}
\label{sec:motivation:entropy}
Let $A_t$ be the agent role at step $t$. We compute the relative
conditional entropy reduction
$R = 1 - H(A_{t+1}\mid A_t) / H(A_{t+1})$.
$R=0$ means the next agent is uniformly random given the current
one; $R=1$ means it is fully determined. Figure~\ref{fig:motivation_locality}(b)
shows $R \in [0.40, 0.48]$ on every workload. Even the lowest-$R$
case eliminates 40\% of the next-step uncertainty conditional on
the current agent, well above what session-level recency alone
extracts. A first-order Markov chain over the agent alphabet
captures most of the available structure; higher-order or
per-prompt models are not required.


\paragraph{\textbf{Observation 3:} transition probabilities vary across
workloads.}
\label{sec:motivation:variability}
If the same framework produced the same transition matrix
regardless of task source, that matrix could be declared once at
deploy time. However, the same six-agent framework, fed
different task sources, produces visibly different routing
patterns, with swings of 27 to 47\,pp on individual edges (e.g.,
Tester$\to$Planner ranges from 0.53 on GSM8K to 1.00 on MT-Bench).
A statically declared graph either over-protects under-used edges
on one workload or under-protects heavy edges on another, so the
transition matrix \emph{must} be learned online from live
dispatch history.

\subsection{Transition Learning}
\label{sec:casestudy:learn}
\label{sec:design:learn}

\noindent\textbf{Agent identity.}
Every block carries a stable \emph{agent identity}
$a \in \mathcal{A}$, computed once at first touch as a content
hash of the first few block hashes after the chat-template prefix
(defaults: \texttt{skip}=4, \texttt{take}=4 for
Llama-3.1-8B-Instruct). The hash lands in the system\,+\,tools
region (the agent anchor measured by $\phiwide$), so two requests
sharing that anchor produce the same identity even when their
session content differs. Block hashes remain the storage key;
agent identity is the prediction key the runtime adds on top.

\noindent\textbf{Markov estimate.}
\textsc{Observe} maintains pairwise counts $n(a', a)$ over a
sliding window; the maximum-likelihood transition is
$\widehat{P}(b \mid a) = n(a, b) / n(a)$, computable in $O(1)$.
The first-order chain is justified by the entropy reduction
$R \in [0.40, 0.48]$ from \S\ref{sec:casestudy:obs}. State is
$O(|\mathcal{A}|^2)$ machine words; on our workloads
$|\mathcal{A}|$ stabilizes below 50, giving $\le$\,20\,KB per
workload.

\subsection{Survival-Probability Eviction}
\label{sec:casestudy:evict}
\label{sec:design:evict}

\noindent\textbf{Tractable proxy.}
We want to score each block by the probability its agent will
fire within the next $K$ steps under $\widehat{P}$. Computing
that exactly costs $O(|\mathcal{A}|^3 K)$ per agent, which is
too expensive on every eviction. We approximate with a hop-count
proxy on a thresholded DAG $G=(\mathcal{A},\mathcal{E})$ with
edge $(a,b)\in\mathcal{E}$ iff $\widehat{P}(b\mid a)\ge\tau$
($\tau=0.01$ preserves $\ge\!99\%$ of observed transitions). A
single BFS from $a_0$ gives a hop count $E[a]$ per agent. Define
\begin{equation}
\tilde p_{\text{surv}}(a) = 1 - \min(E[a],E_{\max})/E_{\max}\in[0,1],
\label{eq:psurv-proxy}
\end{equation}
monotone in $p_{\text{surv}}$ and updated only on agent change.

\noindent\textbf{Block score.}
For block $b$ with agent identity $a_b$ and a normalized recency
residual $\rho_b\in[0,1]$ that reuses the engine's existing
recency clock,
\begin{equation}
\text{score}(b) = w_{\text{pred}}\cdot\tilde p_{\text{surv}}(a_b) + \rho_b,
\quad w_{\text{pred}}=1.0,
\label{eq:score}
\end{equation}
and the engine evicts the lowest-score block. \emph{Predictive
lift}: imminent agents get near-unit additive protection above
recency, so their identity blocks survive cross-session pressure.
\emph{Graceful fallback}: with no learnable structure,
$\widehat{P}$ has high marginal entropy, the DAG fragments, every
$\tilde p_{\text{surv}}$ collapses toward 0, and
score$(b)\to\rho_b$, exactly the LRU clock.

\subsection{Cross-Session Prefetch}
\label{sec:casestudy:prefetch}
\label{sec:design:prefetch}

Eviction protects identity blocks that are \emph{already cached};
prefetching populates identity blocks that are not. Between
scheduler steps, the runtime queries
$a^* = \arg\max_b \widehat{P}(b \mid a_t)$ and issues a
\textbf{warmup}: a request carrying $a^*$'s
system prompt, a one-character user message, and a single
generated token. The warmup populates $a^*$'s identity blocks
(the high-$\phiwide$ region of the anchor) without injecting
session-specific content into the cache. It flows through the
engine's normal serving path, so no kernel or scheduler change is
needed.

\section{Evaluation}
\label{sec:eval}

\begin{figure*}[t]
  \centering
  \begin{subfigure}[b]{0.32\textwidth}
    \includegraphics[width=\textwidth]{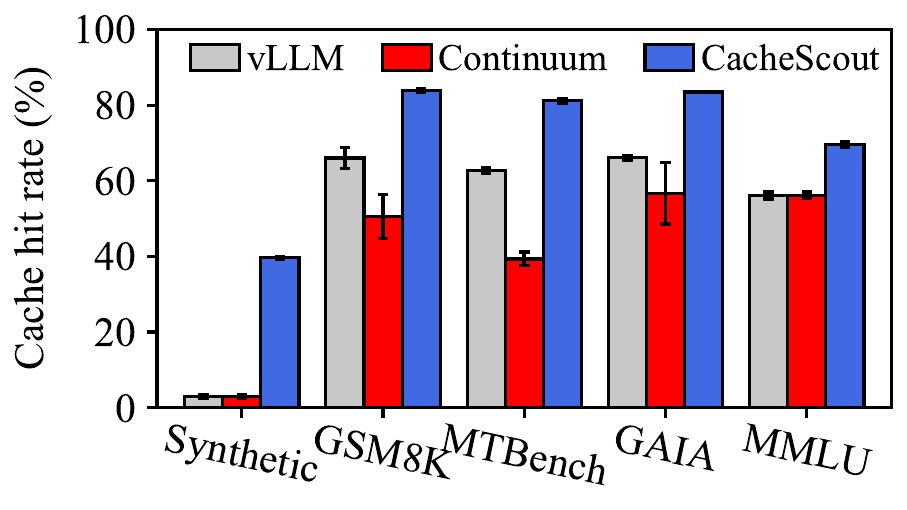}
    \caption{GPU KV cache hit rate.}
    \label{fig:hit_rate}
  \end{subfigure}
  \hfill
  \begin{subfigure}[b]{0.32\textwidth}
    \includegraphics[width=\textwidth]{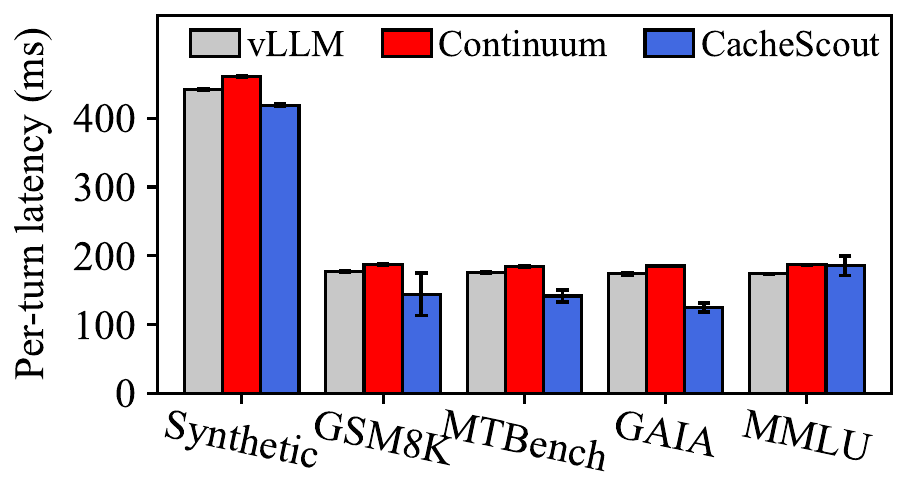}
    \caption{Per-turn end-to-end latency.}
    \label{fig:e2e_latency}
  \end{subfigure}
  \hfill
  \begin{subfigure}[b]{0.32\textwidth}
    \includegraphics[width=0.98\textwidth]{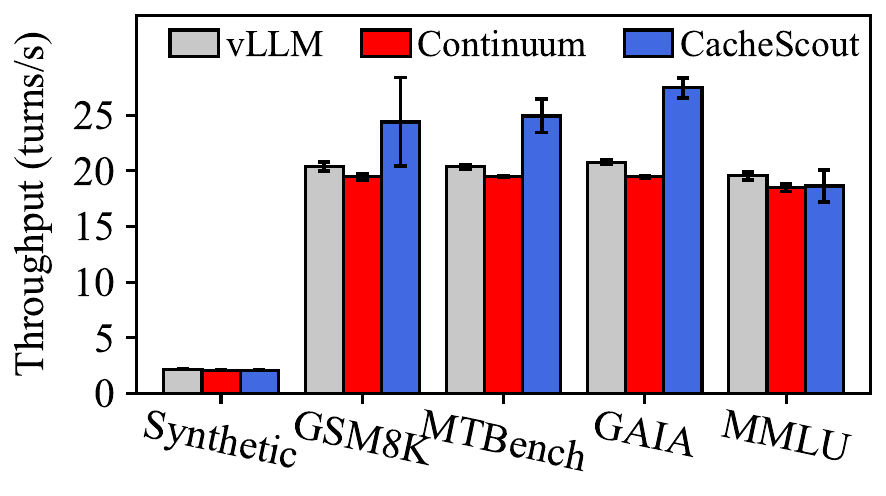}
    \caption{Throughput (turns/s).}
    \label{fig:throughput}
  \end{subfigure}
  \caption{Headline evaluation across five multi-agent workloads,
  comparing \vllm, \continuum, and \sys.
  \textbf{(a)} Cache hit rate; \sys wins 5/5 with +13 to +37\,pp
  lifts.
  \textbf{(b)} End-to-end per-turn latency; \sys drops 6\% to 26\%.
  \textbf{(c)} Throughput; \sys raises 6\% to 14\%.}
  \Description{Three side-by-side bar charts across five workloads
  comparing vLLM, Continuum, and CacheScout on cache hit rate,
  end-to-end latency, and throughput. CacheScout outperforms both
  baselines on hit rate and latency on every workload.}
  \label{fig:evals}
\end{figure*}

\paragraph{Setup.}
Experiments are conducted on NVIDIA H100 80\,GB GPU using Llama-3.1-8B-Instruct~\cite{grattafiori2024llama} as the primary model.
four real-task multi-agent workloads from \S\ref{sec:motivation}
(MMLU, MT-Bench, GAIA, GSM8K) on the same six-agent supervisor
framework with tools and AutoGen's
SelectorGroupChat~\cite{wu2024autogen}, plus a 12-agent
\textsc{Synthetic} chain as a high-structure stress case. The
synthetic runs at concurrency~1 with a 250-block budget; the four
real-task workloads run at concurrency~4 with a 120-block
budget, so the cache is genuinely under pressure. Baselines:
\vllm and \continuum~\cite{li2025continuum}.
50-task traces per workload (948 to 1417 turns). Hit rate is the
cached token fraction (\texttt{cached\_tokens} /
\texttt{prompt\_tokens}).

\paragraph{Headline results.}
Figure~\ref{fig:evals} reports hit rate, end-to-end latency, and
throughput. \sys wins 5/5 over both baselines, lifting hit rate by
+13 to +37\,pp relative to vLLM (the agent anchor is retained
across the session-boundary gap that reactive caches miss),
cutting end-to-end latency by 6\% to 26\%, and raising throughput
by 6\% to 14\%. Continuum's TTL pin reduces hit rate by 9 to
23\,pp on three of the four real-task workloads, a known
trade-off when the inter-turn timescale is shorter than the pin's
horizon.

\paragraph{Cost.}
Hot-path overhead is negligible: \textsc{Observe} runs in
$\sim$1\,$\mu$s mean per block-touch, the BFS-based scorer in
single-digit $\mu$s, and coordinator state stays under 25\,KB at
24 agents. A full accounting (memory and latency microbenchmarks,
sensitivity to cache budget and concurrency, and behavior with
the runtime disabled) is deferred to the extended version.

\section{Related Work}
\label{sec:related}

\paragraph{Point fixes of the same architectural gap.}
A growing line of work inserts agent-aware policy into one
neighbor of the serving stack to address a single row of
Table~\ref{tab:policies}; we view each as a point fix of the same
architectural gap. For KV caching (row 1),
\kvflow~\cite{pan2026kvflow} declares an agent step graph via
SGLang annotations and scores eviction by step distance;
\continuum~\cite{li2025continuum},
Tokencake~\cite{bian2025tokencake}, and
InferCept~\cite{abhyankar2024infercept} pin or migrate KV around
tool calls; LMCache~\cite{liu2025lmcache} provides
an enterprise-scale KV cache layer with cross-tier storage. For
scheduling and placement (rows 4
and 5), Autellix~\cite{luo2025autellix} and
Agent.xpu~\cite{wei2025agent} push agent-aware scheduling into
the engine; LAMPS~\cite{shahout2026fast} schedules augmented
requests by predicted memory footprint;
Parrot~\cite{lin2024parrot} uses developer-supplied Semantic
Variables to colocate dependent requests. For speculative agent
execution (row 6), PASTE~\cite{sui2026act} and
Sherlock~\cite{ro2025sherlock} speculatively execute the
predicted next agent in parallel with the current decode. For
decode adapters and drafters (row 7), multi-LoRA serving
infrastructure swaps adapters at request granularity, and the
speculative-decoding literature swaps drafts per workload but
rarely per agent. For fairness across agents (row 8),
recent work~\cite{zhang2025enabling} enforces fair-share allocation
across multi-modal and multi-agent applications inside the engine
scheduler. Each line of work succeeds at its row, but each
requires deploy-time graphs, hand-coded adapter ids, or per-policy
engine surgery. The runtime layer learns the
structure online and amortizes the same interventions across
rows on a shared \texttt{observe} stream.

\paragraph{Multi-agent frameworks.}
AutoGen~\cite{wu2024autogen}, MetaGPT~\cite{hong2024metagpt},
ReAct~\cite{yao2022react}, and OpenAI Swarm~\cite{openai2024swarm}
dispatch above the engine; SGLang~\cite{zheng2024sglang} sits
between as a deploy-time-declared DSL. Each emits agent identity
but none observe engine events at the granularity the layer
needs.

\paragraph{KV Cache optimizations.}
Non-prefix sharing~\cite{yao2025cacheblend, gim2024prompt}, cross-LLM
reuse~\cite{liu2024droidspeak}, KV-aware
routing~\cite{srivatsa2025preble},
sparse-attention and multimodal
KV~\cite{tu2025vl,
lee2024infinigen},
KV-cache compression and joint
eviction~\cite{ye2026kvcomm, feng2025evicpress}, and structured
generation~\cite{dong2025xgrammar, willard2023efficient} optimize
axes orthogonal to ours and compose cleanly with the layer.

\section{Conclusion}
\label{sec:conclusion}

Multi-agent LLM workloads expose a class of cross-cutting policies
that neither the agent framework nor the serving engine alone can
solve cleanly. We have argued that the right response is
architectural: insert an \emph{agent runtime layer} between the
two tiers, exposing four primitives (\texttt{observe},
\texttt{score}, \texttt{predict}, \texttt{act}) into which any
agent-aware policy plugs, with agent identity as the shared
coordinate. The interface, not any specific policy, is the
contribution; the case study validates that the gap matters in
practice on the one row of Table~\ref{tab:policies} with the
largest immediate serving-cost lever. The other eight rows of
Table~\ref{tab:policies}, including tool-result memoization, batch shaping,
disaggregated P/D placement, speculative agent execution, decode adapters,
fairness, safety, and failure hedging, remain open challenges for future work.

\balance
\bibliographystyle{ACM-Reference-Format}
\bibliography{references}

\end{document}